\def\year{2022}\relax
\documentclass[letterpaper]{article} % DO NOT CHANGE THIS
\usepackage{aaai22}  % DO NOT CHANGE THIS
\usepackage{times}  % DO NOT CHANGE THIS
\usepackage{helvet}  % DO NOT CHANGE THIS
\usepackage{courier}  % DO NOT CHANGE THIS
\usepackage[hyphens]{url}  % DO NOT CHANGE THIS
\usepackage{graphicx} % DO NOT CHANGE THIS
\usepackage{natbib}  % DO NOT CHANGE THIS AND DO NOT ADD ANY OPTIONS TO IT
\usepackage{caption} % DO NOT CHANGE THIS AND DO NOT ADD ANY OPTIONS TO IT
\DeclareCaptionStyle{ruled}{labelfont=normalfont,labelsep=colon,strut=off} % DO NOT CHANGE THIS
\usepackage{algorithm}
\usepackage{algorithmic}

\usepackage{graphicx}
\usepackage{multirow}
\usepackage{multicol}
\usepackage{amsmath}
\usepackage{bm}
\usepackage{caption}
\usepackage{subcaption}
\usepackage{booktabs}
\usepackage{amsthm,amsmath,amssymb}
\usepackage{mathrsfs}

\usepackage{newfloat}
\usepackage{listings}
\floatstyle{ruled}
\newfloat{listing}{tb}{lst}{}
\floatname{listing}{Listing}
\title{Vision Pair Learning: An Efficient Training Framework for Image Classification}
\author {
    Bei Tong,\textsuperscript{\rm 1}
    Xiaoyuan Yu \textsuperscript{\rm 1}
}
\affiliations {
    \textsuperscript{\rm 1} Huawei Cloud\\
    tongbei@huawei.com, yuxiaoyuan@huawei.com
}

\begin{document}

\maketitle

\begin{abstract}
Transformer is a potentially powerful architecture for vision tasks. Although equipped with more parameters and attention mechanism, its performance is not as dominant as CNN currently. CNN is usually computationally cheaper and still the leading competitor in various vision tasks. One research direction is to adopt the successful ideas of CNN and improve transformer, but it often relies on elaborated and heuristic network design. Observing that transformer and CNN are complementary in representation learning and convergence speed, we propose an efficient training framework called Vision Pair Learning (VPL) for image classification task. VPL builds up a network composed of a transformer branch, a CNN branch and pair learning module. With multi-stage training strategy, VPL enables the branches to learn from their partners during the appropriate stage of the training process, and makes them both achieve better performance with less time cost. Without external data, VPL promotes the top-$1$ accuracy of ViT-Base and ResNet-50 on the ImageNet-1k validation set to $83.47\%$ and $79.61\%$ respectively. Experiments on other datasets of various domains prove the efficacy of VPL and suggest that transformer performs better when paired with the differently structured CNN in VPL. we also analyze the importance of components through ablation study.
\end{abstract}

\section{Introduction}

Transformer architecture has attracted much attention since its first application in machine translation, and it has proven effective in various tasks of natural language processing. As a result, the transformer is introduced to computer vision recently. Although researchers have successfully adapted the  transformer for CV tasks, currently CNN based models are superior to transformer based ones in terms of performance and speed. Main reasons include 1) The special structure of transformer lets it pay more attention to global features but ignore detailed information. Because of that, pure transformer-based methods often fail in fine grained tasks such as small object detection. 2) Problems also arise in model training. Due to the relatively large number of parameters, training transformer tends to be more computational expensive than CNN models. For example, comparing with ViT-L/16 \cite{dosovitskiy2020image}, training EfficientNetV2 \cite{tan2021efficientnetv2} is $5\times$-$11\times$ faster. This offers unfriendly hardware requirement and slows down prototyping experiments.

To tackle these problems, effective components of CNN are combined with transformer, for both enhancing performance and decreasing computation cost. These approaches generally fall into two categories, 1) One may utilize CNN blocks to minimize the information loss brought by image tokenization, and reduce the dimension of input feature, so as to balance the performance and speed, e.g. Visual Transformers \cite{wu2020visual}, T2T-ViT \cite{yuan2021tokens}. 2) Modify the general design of network. TNT \cite{han2021transformer} follows the idea of ``deep-narrow'' and makes the transformer deeper and slimmer. CvT \cite{wu2021cvt} replaces linear projection with convolution projection for less computation. Other widely-used techniques, e.g. pruning \cite{zhu2021visual} and sliding window \cite{liu2021swin}, are adopted as well.

\begin{figure*}
  \centering
  \includegraphics[width=0.8\textwidth]{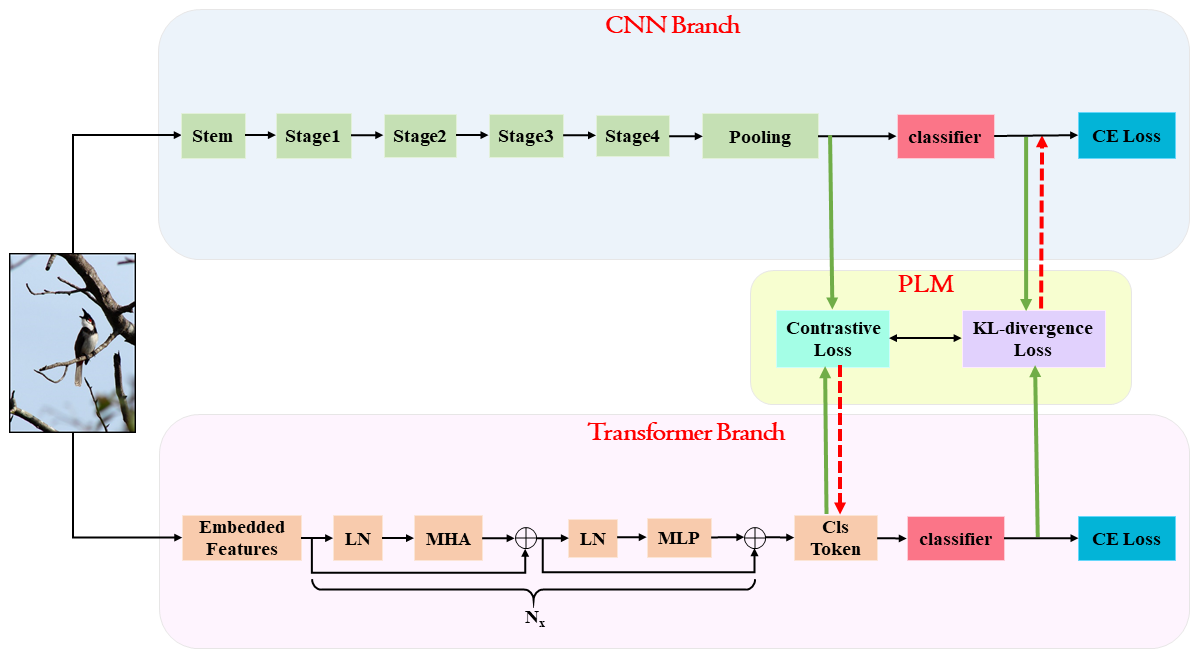}
  \caption{Architecture of Vision Pair Learning. The green bold arrow lines represent the forward pass in PLM. The red dash arrow lines indicate the direction of restricted gradient flow.}
  \label{structure}
\end{figure*}

The performance of above methods are satisfactory, but mixing the advantages of CNN and transformer often relies on elaborated and heuristic network design, and may hurt the transferability. In this paper, we propose a pair learning method based on CNN and transformer for image classification tasks. The method enables two different networks to learn from each other and achieve better performance together. The structure is shown in Figure \ref{structure}.

The network consists of three parts, CNN branch, transformer branch and Pair Learning Module (PLM). Since the target of VPL is to build a high-level training framework for better representation learning, we directly use CNN and transformer baseline models as branches without modification. Being the core part of the model, PLM introduces a contrastive-learning-like task. The most common idea of contrastive learning \cite{he2020momentum,chen2020simple,grill2020bootstrap,chen2020exploring,caron2020unsupervised} is to feed a pair of randomly transformed images to two parameter-sharing branches and predict whether they are from the same source image by the extracted embeddings. Different from that, the branches of our method read the same image as input but are distinct from each other in terms of basic network block. Due to the natural difference of the two branches' network architecture and capacity, the primary model collapse problem in contrastive learning is easily avoided. Meanwhile due to the different inductive biases of global attention and local convolution, the transformer and CNN branches produce representations of diverse views. And the task forces the high-level representation learned by two branches to interact, thus providing extra supervision signals and accelerating the convergence. A distillation-like loss is utilized as well.

Considering that the two branches are of different data efficiency, we additional adjust the training strategy so that the branches can choose to learn from partners during the appropriate stage of the training process. The heterogeneous model with PLM and modified training protocol together constitute vision pair learning. After training, PLM is removed and branches can be used for inference without extra computation.

To summarize, our contributions are three-fold:
\begin{itemize}
\item We propose a new training framework named vision pair learning. By building a heterogeneous model composed of CNN and transformer, and introducing pair learning module (PLM) and multi-stage training strategy, the framework offers both branches faster convergence speed and superior performance in image classification task. After training, PLM is removed and CNN and transformer work without additional component, therefore vision pair learning brings no extra computation burden for inference.
\item With the help of pair learning, ViT-Base obtains $81\%$ top-1 accuracy on ImageNet validation set \cite{deng2009imagenet} in 100-epoch training, which surpasses our reimplemented $77.1\%$ baseline \cite{touvron2020training}. Furthermore, with 300-epoch training, ViT-Base can reach $83.47\%$ and exceeds the official experimental result $81.8\%$ by $1.67$ percent. And ResNet-50 achieve competitive result $79.61\%$. In addition, RegNet-16GF \cite{radosavovic2020designing} and ViT-Base are trained jointly as well for further confirming the effectiveness of the proposed method. The accuracy is increased by $2.36\%$ and $2\%$ respectively compared to independent training.
\item Experiments on more image classification datasets suggest that vision pair learning is a robust and efficient training framework. And ablation studies prove the effectivity of proposed component including simultaneous training, restricted gradient flow, the necessity of different two branches, and so on.
\end{itemize}

\iffalse
(Revision Needed) In summary, we believe that the evidence, challenges, and open questions in this study are worth knowing, if self-supervised Transformers will close the gap in pre-training between vision and language. We hope our data points and experience will be useful to push this frontier.
\fi

\section{Related Work}
\label{gen_inst}

Transformer \cite{vaswani2017attention} and its applications \cite{devlin2018bert,brown2020language} have achieved great success in various natural language processing tasks in recent years. Different from traditional RNN model, its attention-only building block is effective in modeling long-range correlation of input sequences. And large number of parameters introduced by transformer architecture increase the model capacity, as well as the difficulty of optimization. When optimizing a big model, abundant training data is usually necessary for a satisfactory performance. To better apply transformer to data-scarce tasks, NLP researchers propose to utilize the pretraining technique. A transformer is typically pretrained in unsupervised manner with large-scale corpus, and then fine-tuned in downstream tasks. For examples, BERT \cite{devlin2018bert} exploits the encoder part of transformer for sequence modeling and propose two unsupervised pretraining tasks, masked language model and next sentence prediction. BERT is proven powerful and reaches the state-of-the-art in $11$ downstream NLP tasks. Transformer decoder based GPT series models \cite{radford2018improving,radford2019language,brown2020language}, which lead the trend of expanding model size, suggest that pretrained casual language model is good at few-shot/zero-shot learning and natural language generation.

Due to the amazing performance transformer presents in NLP, researchers attempt to introduce it to various CV tasks. \cite{girdhar2019video} proposes an action transformer model for recognizing and localizing human actions in video clips. The transformer encoder based model combines I3D \cite{carreira2017quo} and RPN \cite{ren2016faster} simultaneously, and makes a significant margin in Atomic Visual Actions dataset \cite{gu2018ava}. DETR \cite{carion2020end} transfers transformer to object detection task. The main idea is to learn a 2D representation of an image using CNN backbone and feed it to the transformer to make the final detection prediction. DETR demonstrates significantly better performance on large objects but fails on small objects. Since the results are not particularly satisfactory, its follow-up UP-DETR \cite{dai2020up} puts forward a random query patch detection method and boosts the performance of DETR with faster convergence and higher precision. IPT \cite{chen2020pre} generates corrupted image pairs from ImageNet \cite{deng2009imagenet} and pretrains transformer on them. By fine-tuning the model in low-level CV tasks such as denoising, super-resolution and deraining, IPT outperforms contemporaneous approaches.

Though much progress transformer has made in CV tasks, the majority of proposed approaches just use transformer as a component. And researchers continue to explore transformer-only solutions. ViT \cite{dosovitskiy2020image} firstly shows that a pure transformer is promising in image classification tasks. Even though its performance is weaker than some CNN-based methods, it can overtake them when pretrained with huge datasets. DeiT \cite{touvron2020training} based on ViT further proves transformer can outperform CNN-based methods trained on the same dataset by optimizing the training strategies. Furthermore, the authors presents a distillation strategy for better performance but same inference speed.

% Except for NLP and CV tasks, transformer is also widely used in multi-modality field. CLIP \cite{radford2021learning} trains a big transformer-based model using a dataset of $400$ million pairs of image and text. And it enables zero-shot transfer to downstream tasks. DALL-E \cite{ramesh2021zero} is a small version of the GPT-3 \cite{brown2020language} and can predict visual patterns with language inputs.

Despite that transformer is promising in CV tasks, it is rarely used in production due to the huge computation resources it requires. VPL aims at reducing the cost by combining both advantages of transformer and CNN. It can accelerate the training speed and achieve comparable results with less training iterations. A similar approach \cite{zhang2018deep} has been proposed before, which pays more attention to promoting CNN series by using distillation iteratively. We will introduce the detailed differences between DML and our approach in next section.

%As a consequence, many works focus on decreasing the computation complexity while keeping comparable performance. T2T-ViT \cite{yuan2021tokens} replaces the backbone of ViT with a deep-narrow structure and reduces the parameter counts and MACs of vanilla ViT significantly. Swin Transformer \cite{liu2021swin} proposes a shifted windowing scheme to limit the scope of multi-head attention to avoid huge computation.

\section{Method}

Given a batch of input images $\bm{X}=\{\bm{x}_{1}, \dots, \bm{x}_{N}\}$, image classification task aims to predict the ground-truth label $\bm{Y}=\{\bm{y}_{1}, \dots, \bm{y}_{N}\}$, where $\bm{y}_{i} \in \mathbb{N}^{C}$ is one-hot vector, $N$ is the batch size and $C$ is the class number. The proposed vision pair learning is composed of two branches, pair learning module and multi-stage training. The two branches are original CNN and transformer baseline models, and the pair learning module provide extra supervision signals for them. And the multi-stage training helps branches learn from their partners.

\subsection{Two Branches}

Both branches are widely-used image classification models, except the backbone of one is CNN and the other is transformer. We denote the extracted image embeddings (the input feature of the last fully-connected layer before softmax) as $\bm{H}^{cnn}=\{\bm{h}^{cnn}_{1}, \dots, \bm{h}^{cnn}_{N}\}$ and $\bm{H}^{trans}=\{\bm{h}^{trans}_{1}, \dots, \bm{h}^{trans}_{N}\}$ where $\bm{H}^{cnn}, \bm{H}^{trans}\in\mathbb{R}^{N \times d}$, and $\bm{h}^{cnn}_{i}, \bm{h}^{trans}_{i}\in\mathbb{R}^{d}$. It is notable that if the embeddings of two branches are of different size, we will add an extra affine layer to make sure their embedding sizes are the same. The logits fed to softmax are denoted as $\bm{Z}^{cnn}=\{\bm{z}^{cnn}_{1}, \dots, \bm{z}^{cnn}_{N}\}$ and $\bm{Z}^{trans}=\{\bm{z}^{trans}_{1}, \dots, \bm{z}^{trans}_{N}\}$ where $\bm{Z}^{cnn}, \bm{Z}^{trans}\in\mathbb{R}^{N \times C}$, and $\bm{z}^{cnn}_{i}, \bm{z}^{trans}_{i}\in\mathbb{R}^{C}$. The branches are trained with ground-truth classification labels and cross entropy loss. Their objective function are

\begin{equation}
\label{cnn}
\mathcal{L}_{CE-cnn} = -\frac{1}{N}\sum_{i=1}^{N}\sum_{j=1}^{C}y_{i,j}log\frac{exp(z^{cnn}_{i,j})}{\sum_{k=1}^{C}exp(z^{cnn}_{i,k})} \\
\end{equation}

\begin{equation}
\label{transformer}
\mathcal{L}_{CE-trans} = -\frac{1}{N}\sum_{i=1}^{N}\sum_{j=1}^{C}y_{i,j}log\frac{exp(z^{trans}_{i,j})}{\sum_{k=1}^{C}exp(z^{trans}_{i,k})}
\end{equation}

\subsection{Pair Learning Module}
\label{Pair Learning Module}

Due to the structural differences of CNN and transformer, they focus on different types of feature. The former is better at extracting local features such as texture, edge, etc. And the latter is more proficient in building global receptive field and features. To combine the advantages of both, we introduce the pair learning module. The pair learning module consists of two objectives: proposed contrastive loss and KL-divergence loss. Different from the majority of contrastive learning methods which extract the embeddings of a pair of randomly transformed images with two parameter-sharing branches and predict whether they are from the same source image, we propose to apply contrastive learning to our network in a different perspective. A group of image are simultaneously fed to two branches and two groups of embeddings are extracted respectively. We ask the model to find out which pair of embeddings represent the same image. For example, the probability of $\bm{h}^{trans}_{i}$ being the counterpart of $\bm{h}^{cnn}_{i}$ is formulated as

\begin{align}\label{CL1}
&P(\bm{h}^{cnn}_{i}, \bm{h}^{trans}_{i}) = \frac{\exp(sim(\bm{h}^{cnn}_{i},\bm{h}^{trans}_{i})/\tau)}{Z}
\end{align}

\begin{equation}
\begin{split}
Z = &\sum_{j=1}^{N}\exp(sim(\bm{h}^{cnn}_{i},\bm{h}^{trans}_{j})/\tau) + \\ &\sum_{k=1,k \neq i}^{N}\exp(sim(\bm{h}^{cnn}_{i},\bm{h}^{cnn}_{k})/\tau)
\end{split}
\end{equation}

which can be regarded as a $(2N-1)$-way classification. The loss function of contrastive learning target is formulated as

\begin{equation}
\label{CL2}
\begin{split}
\mathcal{L}_{CL} = -\frac{1}{2N}(&\sum_{i=1}^{N} \log P(\bm{h}^{cnn}_{i}, \bm{h}^{trans}_{i}) + \\ &\sum_{j=1}^{N} \log P(\bm{h}^{trans}_{j}, \bm{h}^{cnn}_{j}))
\end{split}
\end{equation}

where $\tau$ is the temperature and $sim(\cdot,\cdot)$ represents the dot product. As mentioned above, since the two branches in the network are of different architecture, and have unequal model capacity and representation learning ability, our proposed contrastive loss will not lead the model to degradation easily.

We also adopt KL-divegence based loss for representation interaction between branches. Similar to knowledge distillation, we utilize the classification logits as inputs and minimize the KL-divergence between the predicted probability of two branches, which is

\begin{equation}
\label{KL}
\mathcal{L}_{KL} = \frac{1}{N}\sum_{i=1}^{N} KL(g(\bm{z}^{trans}_{i} / \rho)\ || \ g(\bm{z}^{cnn}_{i} / \rho))
\end{equation}

where $\rho$ is the temperature and $g(\cdot)$ represents the softmax function. 

Preliminary distillation experiments show that if we choose only one type of loss function, CNN will perform better when using KL-divergence loss and transformer prefers contrastive loss. Based on the results, we propose to restrict the gradient flow in PLM. To be more specific, the gradients produced by contrastive loss will propagate back to only transformer branch, while the KL-divergence loss will only affect CNN branch, as shown in Figure \ref{structure}. This intentional design is supported by empirical results in experiment section.

Though distillation shares ideas with our methods, the simultaneous training of branches and restricted gradient flow bring better performance to vision pair learning according to our experiments.

\subsection{Multi-Stage Training}
\label{Multi-Stage Training}

When comparing the learning curve of models in independent training, we observe that ResNet (CNN) converges considerably faster than ViT-Base (transformer) in early stage, but ViT-Base catches up gradually when epoch number increases, and reaches a higher final accuracy. Based on this observation, we propose the multi-stage training strategy so that the outperforming branch will help the other one. 

We split the training process into three stages. The overall loss $\mathcal{L}_{VPL}$ changes in different stages, which is formulated as 

\begin{equation}
\label{PLM}
\begin{split}
&\mathcal{L}_{VPL} = \mathcal{L}_{CE-cnn} + \mathcal{L}_{CE-trans} + \\
&\left\{
    \begin{array}{lr}
        \mathcal{L}_{CL}\qquad \qquad \quad \text{first } \quad x\%\text{ epochs}\\
        \mathcal{L}_{CL} + L_{KL}\quad \text{middle } \  (100-x-y)\%\text{ epochs}\\
        \mathcal{L}_{KL}\qquad \qquad \quad \text{last } \quad y\%\text{ epochs}\\
    \end{array}
\right.
\end{split}
\end{equation}

The hard classification labels and corresponding cross-entropy losses always participate in the training. Meanwhile, CNN will lead the training and feeds additional gradient flow to the transformer in the first $x\%$ epochs. During the middle $(100-x-y)\%$ epochs, CNN and transformer will influence each other by optimizing different types of loss functions. In the last $y\%$ epochs, the transformer plays the role of teacher until the training ends.

The training strategy is rather flexible. If $y$ is set to $100$, the model is similar to knowledge distillation with transformer being the teacher. If $x=y=50$, two branches will be the training leader in different half time. And the training becomes one-stage if $x=y=0$. We search for the best setting and  finally set $x=20$ and $y=20$ to get the best performance by empirical results. For more detail about how different settings will influence the accuracy, see ablation study in experiment section.

Our proposed method is similar with the DML \cite{zhang2018deep} in terms of two-tower structure and distillation-like objective, but they differ in more aspects. First, the motivation of VPL is to leverage the two naturally different types of image encoders, CNN and transformer, to avoid the model collapse problem in contrastive learning, while DML aims to help multiple models learning from each other in equal position, regardless of their architectures. Second, based on the motivation, our pair learning module focuses on feature learning and proposes to use contrastive-learning-style objective, in addition to the distillation-style loss in DML. Third, DML update each model iteratively with their independent loss functions which may cause low data efficiency, while VPL proposes a unified loss function and update both models simultaneously. Fourth, we prove that the transformer model performs better when paired with the differently structured CNN rather than another transformer.

\section{Experiment}

In this section, we first introduce the implementation details. And then we evaluate the performance of VPL on ImageNet classification dataset \cite{deng2009imagenet}. More transfer learning experiments are conducted in various datasets as well, including CIFAR-10 \cite{krizhevsky2009learning}, CIFAR-100 \cite{krizhevsky2009learning}, Oxford Flowers-102 \cite{nilsback2008automated}, and Stanford Cars \cite{krause20133d}. The statistics of the datasets are in Table \ref{datasets}. Last, we analyze the importance of different components of the VPL in ablation study.

\begin{table*}[htb]
  \centering
  \begin{tabular}{l|ccc}
    \toprule
    Dataset     & Train images     & Eval Images     & Classes       \\
    \midrule
    ImageNet \cite{deng2009imagenet} & 1,281,167  & 50,000 & 1000    \\
    CIFAR-10 \cite{krizhevsky2009learning} & 50,000 & 10,000 & 10    \\
    CIFAR-100 \cite{krizhevsky2009learning} & 50,000 & 10,000 & 100  \\
    Flowers-102 \cite{nilsback2008automated} & 2,040 & 6,149 & 102   \\
    Stanford Cars \cite{krause20133d} & 8,144 & 8,041 & 196          \\
    \bottomrule
  \end{tabular}
  \caption{Statistics of image classification datasets}
  \label{datasets}
\end{table*}

\subsection{Implementation Details}
\label{Implementation}

\begin{table*}[htb]
  \centering
  \begin{tabular}{c|l|c|c|c|c|c|c}
    \toprule
    \multirow{2}{*}{Epoch} & \multirow{2}{*}{Framework} & \multicolumn{2}{c|}{ImageNet} & \multicolumn{2}{c|}{Real} & \multicolumn{2}{c}{V2}    \\
    \cmidrule{3-8}
    & & Top-1 & Top-5 & Top-1 & Top-5 & Top-1 & Top-5 \\
    \midrule
    \multirow{5}{*}{100}
    & ResNet-50$^{\times}$ \cite{he2016deep} & 76.14 & 92.90 & 82.69 & 95.50 & 63.07 & 84.62 \\
    & ResNet-50$^{\star}$ & 77.22 & 93.34 & 84.18 & 96.08 & 65.19 & 85.88\\
    & VPL ResNet-50 & \textbf{77.45} & \textbf{93.70} & \textbf{84.44} & \textbf{96.31} & \textbf{65.79} & \textbf{86.54} \\
    \cmidrule{2-8}
    & ViT-Base$^{\star}$ & 77.15 & 93.04 & 82.96 & 95.35 & 64.11 & 85.09 \\
    & VPL ViT-Base & \textbf{80.95} & \textbf{95.57} & \textbf{86.57} & \textbf{97.38} & \textbf{69.89} & \textbf{89.32} \\
    \midrule
    \multirow{4}{*}{300}
    & ResNet-50$^{\star}$ & 79.13 & 94.66 & 85.45 & 96.82 & 67.47 & 87.54 \\
    & VPL ResNet-50 & \textbf{79.61} & \textbf{94.88} & \textbf{85.84} & \textbf{96.99} & \textbf{68.28} & \textbf{88.03} \\
    \cmidrule{2-8}
    & ViT-Base$^{\star}$ & 81.65 & 95.69 & 86.84 & 97.21 & 70.85 & 89.54 \\
    & VPL ViT-Base & \textbf{83.47} & \textbf{96.59} & \textbf{88.24} & \textbf{97.94} & \textbf{73.10} & \textbf{91.20} \\
    \bottomrule
  \end{tabular}
  \caption{Top-1 and Top-5 accuracy of ResNet-50, ViT-Base and VPL version. "$\times$" indicates our implementation using the official training strategies and "$\star$" indicates with the reported better training scheme. For fair comparison, the input size is $224^{2}$ and no extra datasets are used for training or pre-training.}
  \label{Overall performance}
\end{table*}

If no further explanation is provided, we use ResNet-50 \cite{he2016deep} and ViT-Base \cite{dosovitskiy2020image} as CNN and transformer backbone respectively in the following experiments.

Different from the training recipe in \cite{he2016deep}, we use AdamW \cite{loshchilov2017decoupled} as the optimizer and cosine decay schedule \cite{loshchilov2016sgdr} to change the learning rate. The maximum learning rate is tuned for different datasets. We set the maximum learning rate of Vit-Base and ResNet-50 to $0.002$ and $0.005$ in ImageNet task and $0.0001$ and $0.0005$ for transferring to other datasets. The weight decay ratio is set to $0.05$ for both branches and $1e^{-8}$ for fine-tuning tasks. $16$ V100 GPUs are used in ImageNet training task and the training batch size is $1,600$. We introduce Exponential Moving Average (EMA) as well and the decay rate changes with the number of training iterations. The EMA operation benefits both ResNet and ViT. For the image pre-processing, we use AutoAugment \cite{cubuk2019autoaugment} and Random Erasing \cite{zhong2020random} to increase the difficulty of the training and help the network converge better. All reported results are average performances of five runs.

% \begin{equation}
% \label{ema}
% decay\_rate = min(default\_decay, \frac{1.0 + iters}{10.0 + iters})
% \end{equation}

It is worth mentioning that we re-train ResNet-50 using settings above. With 100 and 300 epoches' training, the ResNet-50 can obtain 77.2\% and 79.1\% top-1 accurcay on ImageNet validation set respectively, ahead of the popular baseline result used in \cite{ge2021self,zbontar2021barlow,wang2021contrastive}. By empowering the baseline, we hope to select valuable opponents produced by unified experimental setting and make fair and meaningful comparison in the subsequent experiment.

\subsection{Results on ImageNet series datasets}

\begin{figure*}
  \centering
  \begin{subfigure}[b]{0.4\textwidth}
      \centering
      \includegraphics[width=\textwidth]{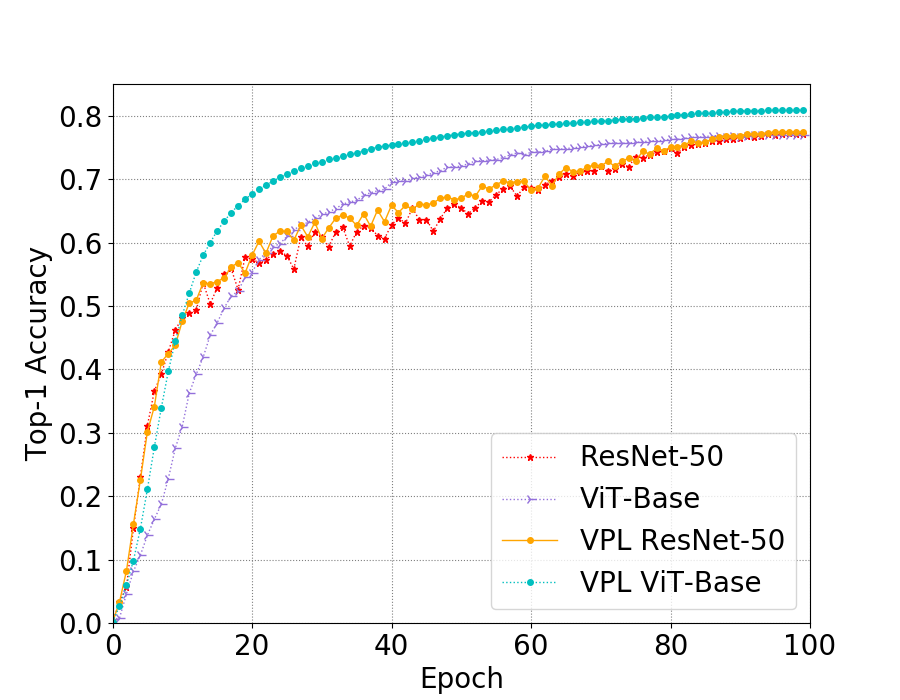}
      \label{converge-100}
  \end{subfigure}
  % \hfill
  \begin{subfigure}[b]{0.4\textwidth}
      \centering
      \includegraphics[width=\textwidth]{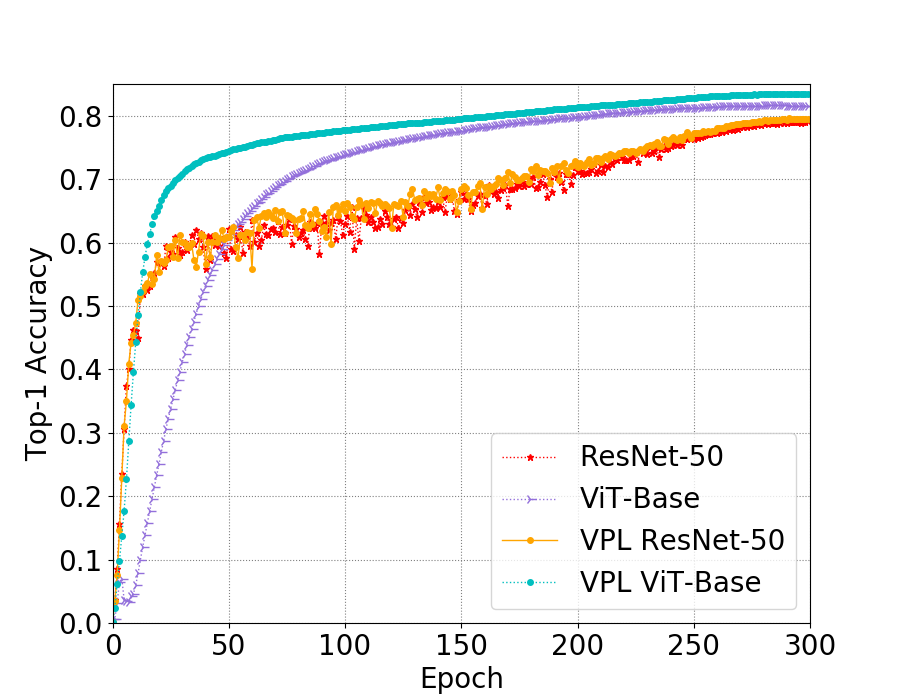}
      \label{converge-300}
  \end{subfigure}
  \caption{Top-1 accuracy v.s. epoch number curves of ResNet-50, ViT-Base and VPL based versions on ImageNet validation set.}
  \label{convergence}
\end{figure*}

Table \ref{Overall performance} reports the overall performance of models on ImageNet \cite{deng2009imagenet}, ImageNet Real \cite{beyer2020we} and ImageNet V2 matched frequency \cite{recht2019imagenet}. We split the experiments into two groups by training epoch number. VPL consistently accelerate the convergence of two branches. For ResNet-50, though our implementation with improved training scheme is stronger than the original version, VPL based ResNet-50 still outperforms them. With $100$-epoch training, VPL push up the top-$1$ accuracy of ViT-Base significantly by $3.5\%$ and achieve a comparable performance to the one trained independently for $300$ epochs. Figure \ref{convergence} shows the convergence curve of models. When the ViT-Base is fully trained with VPL, it reaches a even higher accuracy without any extra supervision signal. The results suggest that the VPL is able to offer two better optimized model with slight modification of network and training strategy.

\begin{table}[htb]
  \centering
  \begin{tabular}{l|c|c|c|c}
    \toprule
    Framework & Epoch & ImageNet & Real & V2\\
    \midrule
    DML ResNet-50 & 300 & 79.00 & 85.61 & 67.37 \\
    DML ViT-Base & 300 & 82.86 & 88.03 & 71.92 \\
    \midrule
    VPL ResNet-50 & 300 & \textbf{79.61} & \textbf{85.84} & \textbf{68.28} \\
    VPL ViT-Base & 300 & \textbf{83.47} & \textbf{88.24} & \textbf{73.10} \\
    \bottomrule
  \end{tabular}
  \caption{Top-1 accuracy of DML and VPL on ImageNet series datasets.}
  \label{DML}
\end{table}

We reimplement the DML \cite{zhang2018deep} and use ResNet-50 and ViT-Base as two small models (i.e. the branches in VPL) in training. The results are shown in Table \ref{DML}. With $300$-epoch training, VPL achieves superior results to DML for both branches. The DML treats all small models equally and let them play the role of student in turn. When the models are of diverse capacity and learning efficiency, letting the smaller model guide the bigger one all the time may not the best choice. Under this situation, VPL utilizes pair learning module and multi-stage training to ensure better performance. Beyond that, DML results in lower data efficiency since it needs do twice iterations for updating both branches for once, which will become more complicated if more networks are included. Experimental results show that VPL is $1.4\times$ faster compared to DML.

We also compare the time cost of VPL and independent training. With $16$ V100 GPUs, independently training ResNet-50 and Vit-Base costs $5$ minutes and $15.33$ minutes for each epoch respectively, and the time cost is $19.58$ minutes for VPL. This suggests that under the circumstances of limited computation resources and need for both CNN and transformer models, VPL is obviously a better choice than independent training in terms of training efficiency and final performance.

\iffalse
\begin{table*}[htb]
  \centering
  \begin{tabular}{c|l|c|c|c|c|c|c}
    \toprule
    \multirow{2}{*}{Epoch} & \multirow{2}{*}{Framework} & \multicolumn{2}{c|}{ImageNet } & \multicolumn{2}{c|}{Real} & \multicolumn{2}{c}{V2}    \\
    \cmidrule{3-8}
    & & Top-1 & Top-5 & Top-1 & Top-5 & Top-1 & Top-5 \\
    \midrule
    \multirow{4}{*}{100}
    & DIS ResNet-50 & 77.18 & 93.48 & 84.11 & 96.14 & 65.45 & 86.10 \\
    & VPL ResNet-50 & \textbf{77.45} & \textbf{93.70} & \textbf{84.44} & \textbf{96.31} & \textbf{65.79} & \textbf{86.54} \\
    \cmidrule{2-8}
    & DIS ViT-Base &\textbf{81.04} & \textbf{95.62} & \textbf{86.89} & \textbf{97.40} & \textbf{70.16} & \textbf{89.32} \\
    & VPL ViT-Base & 80.95 & 95.57 & 86.57 & 97.38 & 69.89 & 89.32 \\
    \midrule
    \multirow{4}{*}{300}
    & DIS ResNet-50 & 79.35 & 94.73 & 85.61 & 96.91 & 67.91 & 87.78 \\
    & VPL ResNet-50 & \textbf{79.61} & \textbf{94.88} & \textbf{85.84} & \textbf{96.99} & \textbf{68.28} & \textbf{88.03} \\
    \cmidrule{2-8}
    & DIS ViT-Base & 83.28 & 96.44 & 88.00 & 97.82 & 72.81 & 90.76  \\
    & VPL ViT-Base & \textbf{83.47} & \textbf{96.59} & \textbf{88.24} & \textbf{97.94} & \textbf{73.10} & \textbf{91.20} \\
    \bottomrule
  \end{tabular}
  \caption{Comparison of VPL and corresponding distillation models on ImageNet. Framework starting with "DIS" means the model is trained under distillation recipe.}
  \label{comparison}
\end{table*}
\fi

\begin{table}[htb]
  \centering
  \begin{tabular}{c|l|c|c|c}
    \toprule
    Epoch & Framework & ImageNet & Real & V2 \\
    \midrule
    \multirow{4}{*}{100}
    & DIS ResNet-50 & 77.18 & 84.11 & 65.45 \\
    & VPL ResNet-50 & \textbf{77.45} & \textbf{84.44} & \textbf{65.79} \\
    \cmidrule{2-5}
    & DIS ViT-Base &\textbf{81.04} & \textbf{86.89} & \textbf{70.16} \\
    & VPL ViT-Base & 80.95 & 86.57 & 69.89 \\
    \midrule
    \multirow{4}{*}{300}
    & DIS ResNet-50 & 79.35 & 85.61 & 67.91 \\
    & VPL ResNet-50 & \textbf{79.61} & \textbf{85.84} & \textbf{68.28} \\
    \cmidrule{2-5}
    & DIS ViT-Base & 83.28 & 88.00 & 72.81 \\
    & VPL ViT-Base & \textbf{83.47} & \textbf{88.24} & \textbf{73.10} \\
    \bottomrule
  \end{tabular}
  \caption{Top-1 accuracy of VPL and corresponding distillation models on ImageNet. Framework starting with "DIS" means the model is trained under distillation recipe.}
  \label{comparison}
\end{table}

\subsection{Transfer Learning}

% \begin{table}[htb]
% \begin{floatrow}
%   \caption{Top-1 and Top-5 accuracy of VPL, ResNet-50 and ViT-Base on CIFAR-10, CIFAR-100, Flowers-102 and Stanford Cars. We train the networks using input size of $224^{2}$ and epoch of 300. "$\star$" depicts using ImageNet pre-training. The results of ViT-Base surpass \cite{touvron2020training} and \cite{dosovitskiy2020image}}.
%   \label{performance on other datasets}
%   \centering
%       \begin{tabular}{l|c|c}
%         \toprule
%         Framework & CIFAR-10 & CIFAR-100 \\
%         \midrule
%         ResNet-50 &  &  \\
%         VPL ResNet-50 & 98.41 & - \\
%         \midrule
%         DeiT-B dist \cite{touvron2020training} & 99.1 & 91.3 \\
%         ViT-Base &  &  \\
%         VPL ViT-Base & 99.10 & - \\
%         \bottomrule
%       \end{tabular}
      
%       \begin{tabular}{l|c|c}
%         \toprule
%         Framework & Flowers & Cars \\
%         \midrule
%         ResNet-50 & 98.23 & 94.07\\
%         VPL ResNet-50 & 98.28 & 94.11 \\
%         \midrule
%         DeiT-B dist \cite{touvron2020training} & 98.8 & 92.9 \\
%         ViT-Base & 98.41 & 92.50 \\
%         VPL ViT-Base & 98.80 & 94.14 \\
%         \bottomrule
%       \end{tabular}
%   \end{floatrow}
% \end{table}

We further investigate the transferability of VPL. We first pretrain the models on ImageNet dataset then transfer them to other datasets for further finetuning. The results shown on Table \ref{transfer} suggest that VPL gives two branches a big boost across these datasets and outperforms independently trained classification models. Compared with ResNet-50, ViT-Base benefits from VPL more and surpasses the original ViT-B/16 \cite{dosovitskiy2020image} significantly on three of four datasets. Furthermore, it also achieve better performance than the distillation version of DeiT-B \cite{touvron2020training} which uses RegNetY-$16$GF \cite{radosavovic2020designing} as a teacher for distillation.

\begin{table}[htb]
  \centering
  \begin{tabular}{l|c|c|c|c}
    \toprule
    Framework & CF-10 & CF-100 & Flowers & Cars\\
    \midrule
    ResNet-50 & 98.27 & 87.51 & 98.23 & 94.07\\
    VPL ResNet-50 & \textbf{98.41} & \textbf{87.87} & \textbf{98.28} & \textbf{94.11} \\
    \midrule
    ViT-B/16 & 98.13 & 87.13 & 89.49 & - \\
    DeiT-B & 99.10 & 91.30 & 98.80 & 92.90\\
    ViT-Base & 99.08 & 91.36 & 98.41 & 92.50\\
    VPL ViT-Base & \textbf{99.10} & \textbf{91.62} & \textbf{98.80} & \textbf{94.14}\\
    \bottomrule
  \end{tabular}
  \caption{Top-1 accuracy of different models and VPL in tranfer learning. CF-10, CF-100, Flowers and Cars stand for CIFAR-10, CIFAR-100, Flower-102 and Stanford Cars datasets respectively.}
  \label{transfer}
\end{table}

\subsection{Ablation Study}
\label{Ablations}

\subsubsection{Different Architecture}

\begin{table}[htb]
  \centering
  \begin{tabular}{l|l|c|c|c}
    \toprule
    Branch 1 & Branch 2 & B1 & B2 & B1 solo \\
    \midrule
    ResNet-50 & ViT-Base & 79.61 & 83.47 & 79.13 \\
    RegNet-16GF & ViT-Base & 83.57 & 83.78 & 81.21 \\
    ViT-Small & ViT-Base & 80.66 & 81.71 & 79.46 \\
    \bottomrule
  \end{tabular}
  \caption{Top-1 accuracy of different architecture setting of VPL on ImageNet dataset. ``B1'' and ``B2'' refers to the accuracy of ``Branch 1'' and ``Branch 2'' in the VPL. ``B1 solo'' refers to the accuracy of ``Branch 1'' in independent training. }
  \label{diff_arch}
\end{table}

To test the compatibility of VPL and investigate how the architectures of branches will influence the performance, we first substitute the ResNet-50 with a stronger and larger CNN, RegNet-16GF. The results are shown in Table \ref{diff_arch}. As expected, both branches get better results than solo training. ViT-Base reaches a higher accuracy with the help of a better companion and RegNet-16GF get boosted as well. This suggests that VPL is relatively insensible to the architecture of CNN branch.

To further explore how much help the architecture difference can bring in VPL, we use ViT-Small whose parameter number is close to ResNet-50, as the backbone of CNN branch, i.e. both branches are ViT based model and share the same inductive bias. Although both transformers reach improved accuracy when comparing with independent training, the performance of ViT-Base is significantly inferior to the one paired with ResNet-50. The results lead to an interesting conclusion that a weaker model may not be a bad teacher if it can offer informative and representations and signals of different views.

\subsubsection{Distillation}
\label{Distillation}
Considering that vision pair learning trains branches with multi-stage strategy and promotes both branches to converge simultaneously, it is different from simple distillation. For convincing results, we develop distillation based models and compare them with VPL models. The distillation scheme is as follows, first one branch (teacher) is trained with ground-truth labels and best parameter setting from scratch (baselines in Table \ref{Overall performance}), then its weights are frozen and we train the other branch (student) with both hard labels and losses defined in pair learning module.

Since the VPL based models are always stronger than the independently trained ones (teacher models) as shown in Table \ref{Overall performance}, we focus on the results of student models in Table \ref{comparison}. Though the gap between two groups of models is close, VPL is able to produce slightly better classification accuracy in most settings. And one-step VPL is relatively easier for implementation than two-step distillation. 

% \iffalse
% \paragraph{Loss Function} We conduct series experiments of different pair learning function and contrastive temperature. It can be find from Table \ref{pair learning loss}, with unsuitable training strategies, the convergence of ResNet can even be misled.

% \begin{table}[htb]
%   \centering
%   \begin{tabular}{l|c|c|c|c|c|c}
%     \toprule
%     \multirow{2}{*}{Framework} & \multicolumn{2}{c|}{ImageNet} & \multicolumn{2}{c|}{Real} & \multicolumn{2}{c}{V2}    \\
%     \cmidrule{2-7}
%     & Top-1 & Top-5 & Top-1 & Top-5 & Top-1 & Top-5 \\
%     \midrule
%     VPL ResNet-50 & \textbf{77.45} & \textbf{93.70} & \textbf{84.44} & \textbf{96.31} & \textbf{65.79} & \textbf{86.54} \\
%     VPL ResNet-50 soft & 75.48 & 92.78 & 82.84 & 95.63 & 63.72 & 84.68 \\
%     VPL ResNet-50 hard & 73.94 & 92.10 & 81.66 & 95.03 & 62.09 & 84.09 \\
%     VPL ResNet-50 adaptive & 76.52 & 93.21 & 83.78 & 95.92 & 64.39 & 85.67 \\
%     \midrule
%     VPL ViT-Base & \textbf{80.95} & \textbf{95.57} & \textbf{86.57} & \textbf{97.38} & \textbf{69.89} & \textbf{89.32} \\
%     VPL ViT-Base soft & 79.74 & 94.83 & 85.53 & 96.69 & 67.66 & 88.02 \\
%     VPL ViT-Base hard & 80.00 & 94.90 & 85.63 & 96.81 & 68.75 & 88.00 \\
%     VPL ViT-Base adaptive & 79.39 & 94.81 & 85.39 & 96.75 & 67.72 & 88.14 \\
%     \bottomrule
%   \end{tabular}
%   \caption{Ablation studies of PLM. The epoch is 100.}
%   \label{pair learning loss}
% \end{table}
% \fi

\subsubsection{Gradient Flow in Pair Learning Module}
\label{gradient flow}

\iffalse
\begin{table*}[htb]
  \centering
  \begin{tabular}{l|c|c|c|c|c|c}
    \toprule
    \multirow{2}{*}{Framework} & \multicolumn{2}{c|}{ImageNet} & \multicolumn{2}{c|}{Real} & \multicolumn{2}{c}{V2}    \\
    \cmidrule{2-7}
    & Top-1 & Top-5 & Top-1 & Top-5 & Top-1 & Top-5 \\
    \midrule
    VPL ResNet-50 & \textbf{79.61} & \textbf{94.88} & \textbf{85.84} & \textbf{96.99} & \textbf{68.28} & \textbf{88.03} \\
    VPL ResNet-50 Bi & 78.63 & 94.40 & 85.32 & 96.69 & 66.83 & 87.46 \\
    \midrule
    VPL ViT-Base & \textbf{83.47} & \textbf{96.59} & \textbf{88.24} & \textbf{97.94} & \textbf{73.10} & \textbf{91.20} \\
    VPL ViT-Base Bi & 82.57 & 96.28 & 88.01 & 97.89 & 72.03 & 90.52 \\
    \bottomrule
  \end{tabular}
  \caption{Comparison between different designs of gradient flow in pair learning module. Models end with "Bi" depicts the loss functions will influence both branches in optimization.}
  \label{loss strategy}
\end{table*}
\fi

\begin{table}[htb]
  \centering
  \begin{tabular}{l|c|c|c}
    \toprule
    Framework & ImageNet & Real & V2 \\
    \midrule
    VPL ResNet-50 & \textbf{79.61} & \textbf{85.84} & \textbf{68.28} \\
    VPL ResNet-50 Bi & 78.63 & 85.32 & 66.83 \\
    \midrule
    VPL ViT-Base & \textbf{83.47} & \textbf{88.24} & \textbf{73.10} \\
    VPL ViT-Base Bi & 82.57 & 88.01 & 72.03 \\
    \bottomrule
  \end{tabular}
  \caption{Comparison of Top-1 accuracy between different designs of gradient flow in pair learning module. Models end with "Bi" depicts the loss functions will influence both branches in optimization.}
  \label{loss strategy}
\end{table}

As described in Pair Learning Module, we control the gradients produced by the proposed two loss functions and let each of them influence only one branch in optimization. To prove the effectivity of this design, we conduct experiments where the restriction is cancelled and the gradient flow will update the weights of both branches. As shown in Table \ref{loss strategy}, the restriction brings consistently better performance for VPL on ImageNet datasets. As for the reason, we think there is no one-size-fit-all loss function in vision pair learning and models of different architecture should choose adequate loss function for better performance.

\subsubsection{Multi-stage training strategies}

\iffalse
\begin{table*}[htb]
  \centering
  \begin{tabular}{l|c|c|c|c|c|c}
    \toprule
    \multirow{2}{*}{Framework} & \multicolumn{2}{c|}{ImageNet} & \multicolumn{2}{c|}{Real} & \multicolumn{2}{c}{V2}    \\
    \cmidrule{2-7}
    & Top-1 & Top-5 & Top-1 & Top-5 & Top-1 & Top-5 \\
    \midrule
    VPL ResNet-50 \ \ $x=y=0$ & 79.51 & 94.81 & 85.75 & 96.99 & 68.06 & 87.90 \\
    VPL ResNet-50 \ \ $x=y=20$ & \textbf{79.61} & \textbf{94.88}  & \textbf{85.84} & \textbf{96.99} & \textbf{68.28}  & \textbf{88.03} \\
    VPL ResNet-50 \ \ $x=y=40$ & 79.53 & 94.82 & 85.76 & 96.86 & 67.72 & 87.94 \\
    \midrule
    VPL ViT-Base \ \ \ \ $x=y=0$ & 83.40 & 96.42 & 88.10 & 97.86 & 72.99 & 90.70 \\
    VPL ViT-Base \ \ \ \ $x=y=20$ & \textbf{83.47} & \textbf{96.59} & \textbf{88.24} & \textbf{97.94} & \textbf{73.10} & \textbf{91.20} \\
    VPL ViT-Base \ \ \ \ $x=y=40$ & 83.04 & 96.36 & 87.74 & 97.62 & 72.46 & 90.68 \\
    \bottomrule
  \end{tabular}
  \caption{Results of multi-stage training of different proportions. The percent number indicates the proportion of the first stage and the last stage. }
  \label{training stage ablation}
\end{table*}
\fi

\begin{table}[htb]
  \centering
  \begin{tabular}{l|c|c|c}
    \toprule
    Framework & ImgNet & Real & V2 \\
    \midrule
    VPL ResNet-50 \ $x=y=0$ & 79.51 & 85.75 & 68.06 \\
    VPL ResNet-50 \ $x=y=20$ & \textbf{79.61} & \textbf{85.84} & \textbf{68.28} \\
    VPL ResNet-50 \ $x=y=40$ & 79.53 & 85.76 & 67.72 \\
    \midrule
    VPL ViT-Base \ \ \ $x=y=0$ & 83.40 & 88.10 & 72.99 \\
    VPL ViT-Base \ \ \ $x=y=20$ & \textbf{83.47} & \textbf{88.24} & \textbf{73.10} \\
    VPL ViT-Base \ \ \ $x=y=40$ & 83.04 & 87.74 & 72.46 \\
    \bottomrule
  \end{tabular}
  \caption{Results of multi-stage training of different proportions. The percent number indicates the proportion of the first stage and the last stage. }
  \label{training stage ablation}
\end{table}

We investigate how the proportions of different stages in training will affect the model performance. To make the comparison brief, we set the proportions of the first stage and the last stage to be equal, i.e. $x=y$ in multi-stage training, and change $x$. In Table \ref{training stage ablation}, the results of different proportions vary in a relatively narrow range. The multi-stage training slightly enhances the accuracy and the improvement is limited. And the best performance is achieved when x and y are equal to $20$.

\section{Conclusion}
In this paper, we have proposed a new training framework named vision pair learning (VPL) for image classification task. The VPL makes use of the advantages of CNN and transformer and promote theirs performances simultaneously. We prove the effectivity of VPL in ImageNet series datasets and four downstream datasets. And the experiments further verify its ability in building strong representation. Future work will be on improving the model’s capability and flexibility to deal with more different structures and verify its ability in other vision fields. We hope the viewpoints and experimental results described in this work will be helpful to the follow-up work.

% Use \bibliography{yourbibfile} instead or the References section will not appear in your paper
\bibliography{aaai22}

\end{document}